\documentclass[runningheads]{llncs}
\usepackage{graphicx}
\usepackage{amsmath}
\usepackage{amssymb}
\usepackage{array}
\usepackage{algorithm}  
\usepackage{algpseudocode}  
\usepackage{subfigure}
\usepackage{color}
\usepackage[misc]{ifsym}
\usepackage{diagbox}
\usepackage{multirow}
\usepackage{color}

\begin{document}
\title{Multi-scale Edge-based U-shape Network for Salient Object Detection}
%
%
\title{Multi-scale Edge-based U-shape Network for Salient Object Detection}

\author{Han Sun\textsuperscript{1(\Letter)}\and
	Yetong Bian\textsuperscript{1} \and Ningzhong Liu\textsuperscript{1} \and
	Huiyu Zhou\textsuperscript{2}}

\authorrunning{H. Sun et al.}

\institute{\textsuperscript{1} Nanjing University of Aeronautics and Astronautics, Jiangsu Nanjing, China\\
	\textsuperscript{2} School of Informatics, University of Leicester, Leicester LE1 7RH, U.K\\
	\email{sunhan@nuaa.edu.cn}	
}
\maketitle              
\begin{abstract}
Deep-learning based salient object detection methods achieve great improvements. However, there are still problems existing in the predictions, such as blurry boundary and  inaccurate location, which is mainly caused by inadequate feature extraction and integration. In this paper, we propose a Multi-scale Edge-based U-shape Network (MEUN) to integrate various features at different scales to achieve better performance. To extract more useful information for boundary prediction, U-shape Edge Network modules are embedded in each decoder units. Besides, the additional down-sampling module alleviates the location inaccuracy. Experimental results on four benchmark datasets demonstrate the validity and reliability of the proposed method. Multi-scale Edge-based U-shape Network also shows its superiority when compared with 15 state-of-the-art salient object detection methods.

\keywords{Salient object detection  \and Multi-scale feature \and Edge-based}
\end{abstract}

\section{Introduction}
Saliency Object Detection (SOD) is a significant branch of computer vision. It is involved in plenty of computer vision tasks, such as video summarization \cite{1An2002}, visual object tracking \cite{3Adaptive2012}, semantic segmentation \cite{5Joint2019}. The salient object is defined as one or more objects that are the most attractive in an image. Saliency Object Detection aims to segment the object with its boundary and background accurately. It can be viewed as a binary classification task to assign foreground pixels to saliency and background pixels to non-saliency.

At the very beginning, SOD models mainly depend on manual features such as texture, color and global contrast \cite{7Saliency2012}.
 Until 2015, the sudden rise of neural network sets off a wave of wind in the SOD and even the whole field of computer vision. Most scholars turned their attention to neural network models with a great capacity to extract multi-level and multi-scale information, especially after the Fully Convolutional Neural Network (FCN) \cite{9Fully2015} was proposed.
 
Although SOD models with convolution neural networks (CNNs) achieved remarkable achievements in recent years, there are still two main problems to be solved. First, predictions around the boundary areas are prone to make mistakes. Besides, the background and foreground have high similarities in some images, making the models confused about object location. These problems are virtually triggered by improper multi-scale feature integration and information loss. 

In the past years, most researches are devoted to improving the border region of salient objects, for example, adding some edge information to the framework \cite{12WFNet2020} \cite{11EGNet2019}, utilizing top local features to refine the saliency map via multiple cycles \cite{13R3Net2018} or raising the weight of edge pixel error punishment in loss function \cite{14Contour2021}. Recently, a few articles start to refocus on the accuracy of the overall positioning of salient objects \cite{15EDN2020}.

Inspired by these articles, we propose a Multi-scale Edge-based U-shape Network (MEUN), which improves the object location and boundary prediction by additional down-sampling and edge complement. As known, the features extracted from shallow layers usually have high resolution with abundant detailed information. If these features could be fully utilized in the network, the prediction performance around the saliency borders will be greatly promoted. The features drawn from the deep layers contain rich global textual clues. However, too many straight down-sampling operations will lead the features to lose detailed information and influence the prediction of boundary areas. If we simply up-sample the down-sampled features to the size of inputs, the prediction must be too coarse to meet the requirements of SOD nowadays. So we design our module with the encoder-decoder structure called “U-shape Network” \cite{10U2Net2020}.The outputs of each unit in the encoder are transmitted to corresponding unit in the decoder to supply some shallow details for the deep global semantic features. Such a solution could alleviate the problems mentioned above. Besides, we designed the U-shape edge network (UEN) block and the additional down-sampling module (ADM) to further explore the fine details at the bottom sides and the semantic information at the top sides, separately.

In general, to improve the accuracy of the object location and the details of the object boundary, our method properly extracts and merges features from the deep layers and the shallow layers. The main contributions are summarized as following three points:
\begin{itemize}
	\item [$\bullet$]We propose the U-shape edge network (UEN) block to fuse the edge information and features extracted by the backbone. Based on the originally advantageous U-shape structure, this module could efficiently add boundary information so that the boundary prediction of saliency maps can be improved.
\end{itemize}
\begin{itemize}
	\item [$\bullet$]An additional down-sampling module (ADM) is designed to further extract useful global structural information. It makes the network proposed going deeper than other networks, so our method can obtain a more accurate saliency object position.
\end{itemize}
\begin{itemize}
	\item [$\bullet$]We build an efficient framework to fully combine and fuse edge information, detailed information and semantic clues. Many experiments are conducted to illustrate the validity of our algorithm and this model could surpass most models on four large-scale salient object detection datasets.
\end{itemize}
\section{Related Work}
Because our method designs a novel multi-scale feature integration U-shape Network based on edge information, this part briefly includes three aspects of works, U-shape models, edge-fused models and multi-scale feature aggregation models.

\textbf{U-Net based.} U-shape network is a variant of Fully Convolutional Networks and is widely used in image segmentation and saliency object detection. This structure has a strong capacity to make the features from different modules interact with each other. It was first proposed in \cite{26UNet2015} for biomedical image segmentation in 2015. Recently, many saliency models adopt U-structure to obtain multi-scale features and efficient aggregation. Zhou et al. \cite{27HUANet2020} design a module based on U-Net, in which an attention mechanism is taken to jointly enhance the quality of salient masks and reduce the consumption of memory resources. Qin et al. \cite{10U2Net2020} design a two-level nested U-structure without using any pre-trained backbones from image classification. Although these models take advantage of the U-structure, there are still some limitations. For example, U2Net \cite{10U2Net2020}  is extremely complex with a large number of parameters up to 176.3M.

\textbf{Edge based.} Boundary prediction accuracy is always a problem that most models cannot deal with well. In recent years, edge information is found to be a valuable complement to the salient boundary prediction. EGNet \cite{11EGNet2019} is design to explicitly model complementary salient object information and salient edge information within the network to preserve the salient object boundaries. To supplement semantics and make networks focus on object boundaries, Cen et al. \cite{12WFNet2020} introduce an edge-region complementary strategy and an edge-focused loss function to predict salient maps with clear boundaries accurately. Li et al. \cite{30HFFNet2020} propose an edge information-guided hierarchical feature fusion network.
Zhou et al. \cite{29Hierarchical2021} designed a multi-stage and dual-path network to jointly learn the salient edges and regions, in which the region branch network and edge branch network can interactively learn from each other. These methods improve the boundary prediction of the salient object to some extent, but the overall positioning lacks novel improvement.

\textbf{Multi-scale feature aggregation based.} The problems mentioned above are also partly caused by inappropriate aggregation and fusion. Thus, many articles looked for more efficient and effective methods to aggregate low-level features with detailed information and high-level features with semantic clues. Feng et al. \cite{31Residual2020} employ a Dilated Convolutional Pyramid Pooling (DCPP) module to generate a coarse prediction based on global contextual knowledge. In MINet \cite{32MINet2020}, the aggregate interaction module can efficiently utilize the features from adjacent layers through mutual learning, while the self-interaction module makes the network adaptively extract multi-scale information from data and better deal with scale variation. Li et al. \cite{33Learning2020} present a residual refinement network with semantic context features, including a global attention module, a multi-receptive block module, and a recurrent residual module. These methods are mainly innovative in feature extraction and fusion, but they do not explore features with more possibilities.

\section{Methodology}
In this section, the proposed model is discussed in detail. The whole method is further separated into four parts. In the first part, we give an overview of the network architecture. Then we introduce the U-shape edge network block and the additional down-sampling module. Finally, the use of loss function is explained clearly.

\subsection{Overview}
\begin{figure*}[ht]
	\centering
	\includegraphics[width=12cm]{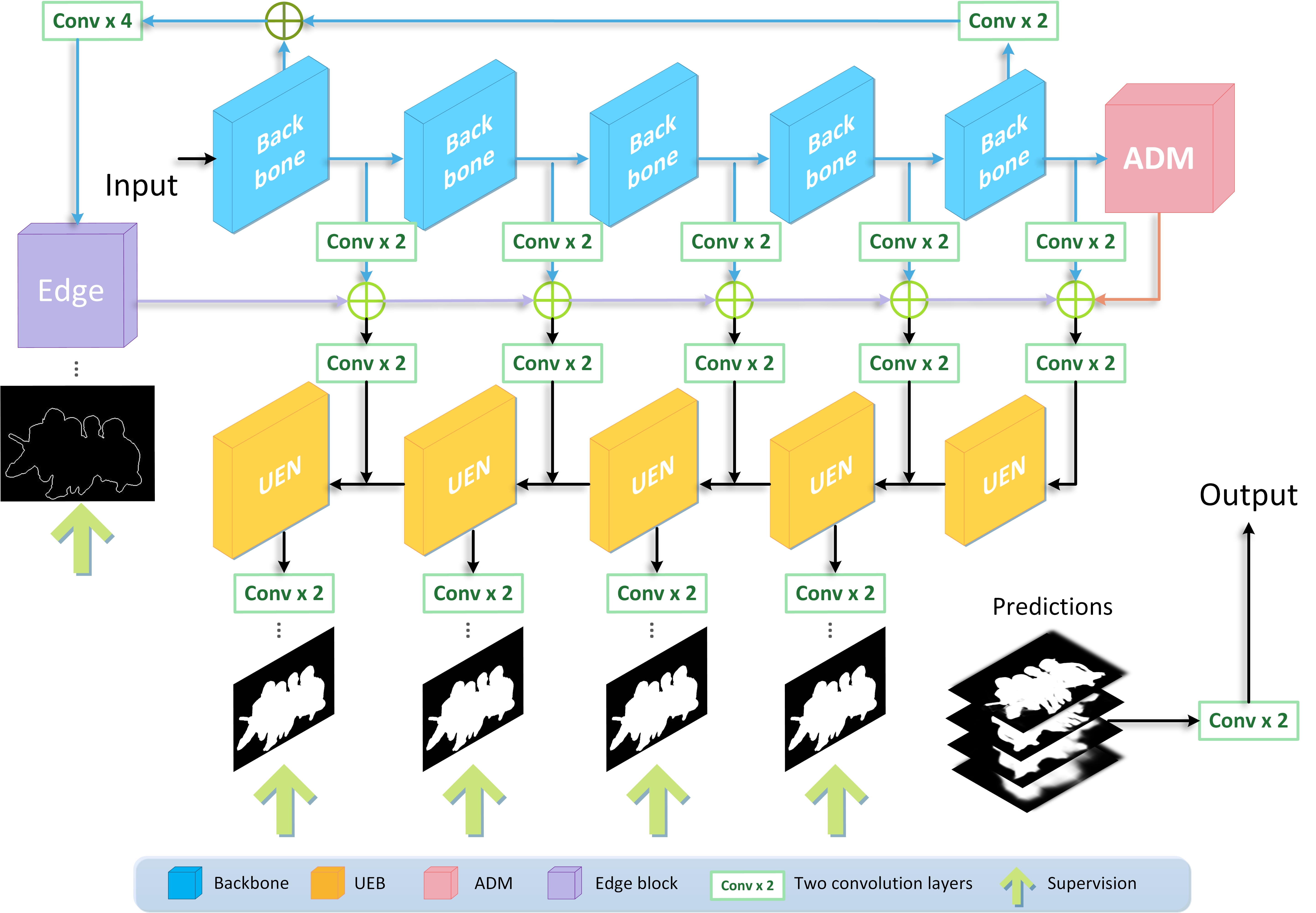}
	\caption{The overview of the architecture.}
	\label{fig1}
\end{figure*}

Fig. \ref{fig1} presents the overall framework, mainly containing the UENs and ADM. The encoder could adopt any common backbone as its encoder, such as ResNet50 \cite{22Deep2016}. The backbone is divided into five groups according to the size of output features. One group's output is transmitted to the next group and the module with the corresponding scale in the decoder. 
We suppose that detailed supplement is not enough for promoting edge prediction, so edge information is introduced as a complement for better performance. Firstly, it should be clear that what kind of edge is needed.
 The background prediction will be disturbed by the background edges marked as saliency. Because in a saliency map, background pixels should be marked as non-salient. To avoid the disturbance from background edges, a salient edge ground truth is used to supervise the edge map generation.
  The edge features are down-sampled to different scales of the branches. Then processed edge feature is inputted into the UENs.

We also design an Additional Down-sampling Module (ADM) to mine more semantic clues with larger and wider receptive field, which can help locate accurately and precisely.
 We build the MEUNetwork based on ResNet-50. The final output of ResNet-50 is at $7 \times 7$ scale, so ADM down-samples the input feature map by a factor of two. Then, the output of ADM is transmitted to the decoder along with the features from the backbone.
The decoder consists of five U-shape edge networks (UEN), and the UENs are connected from the bottom up. The input of the module at the bottom is the element-wise addition of $Feature_{ADM}$ and $Feature_{b5}$. It can be formulated as
\begin{equation}
	\operatorname{Input}_{5}=\operatorname{Conv}_{-}2_{3\times3}\left(\operatorname{Conv}_{-}2_{3\times3}\left(\text {Feature}_{b5}\right)+\text {Feature}_{ADM}\right),
\end{equation}
in which $Input_{5}$ denotes the input of the UEN at the bottom. $Conv_{-}2_{3\times3}$ is two sets of one $3\times3$ convolutional layer followed by the batch normalization and ReLU activation, which is used as the basic convolutional unit in this network. $Feature_{ADM}$ and $Feature_{b5}$ are the output of fifth group filters in the encoder and ADM, respectively. 
The input of other UEN blocks can be written as 
\begin{equation}
	Input_{i}=Conv_{-} 2_{3\times3}\left(Conv_{-}2_{3\times3}\left(Feature_{b_{i}}\right)+ Edge_{sal}\right), \quad i=1,2,3,4
\end{equation}
 $Input_{i}$ is the input of the  $i$th UEN block. $Feature_{b_{i}}$ represents the output of the $i$th block in the encoder and $Edge_{sal}$ denotes the edge supplement. It needs to mention that the outputs with different channels from the backbone are squeezed to 128 channels by a $1\times1$ convolution for the reduction of parameters. Besides, to achieve the best balance between efficiency and quality, edge feature and features in ADM and UENs are all set as 128 channels.
 
Besides being sent to the next module, the output of each UEN is used to generate a saliency map through two $3\times3$ convolution layers. Each intermediate prediction is up-sampled to the size of the original image for supervision. In the decoder, the deepest features at $7\times7$ scale are restored step by step. Compared with other predictions, the last prediction with the highest resolution is the closest to what we want. The final output of the model is the fusion of all the intermediate predictions integrated by convolutional layers. Of course, this step could be skipped because the last saliency map of the decoder is good enough to be the final result.

\subsection{UEN module}

\begin{figure}[htp!]
	\centering
	\subfigure[UEN\_5]{
		\centering
		\includegraphics[width=0.47\textwidth]{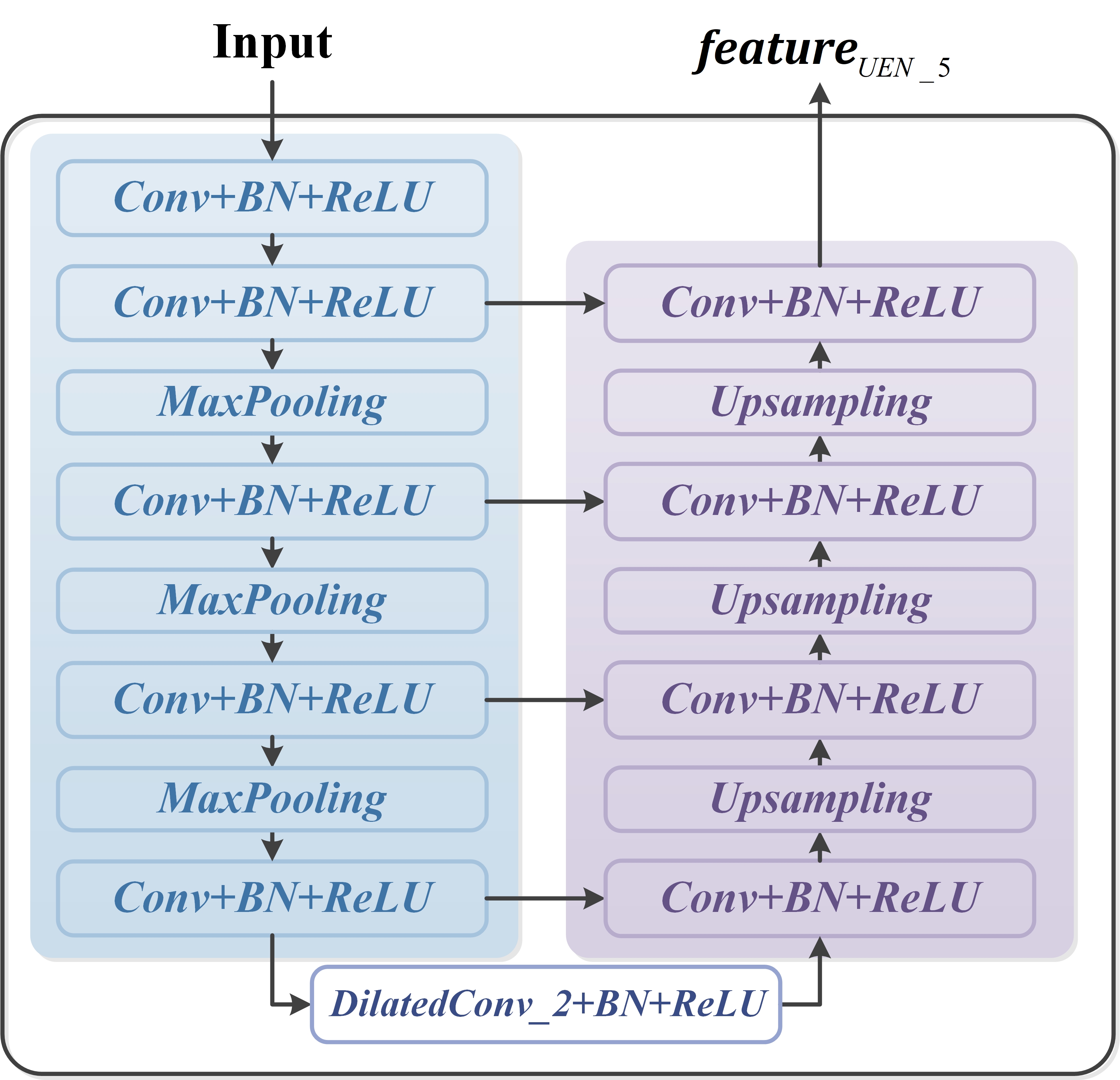}
	}
	\subfigure[UEN\_A]{
		\centering
		\includegraphics[width=0.47\textwidth]{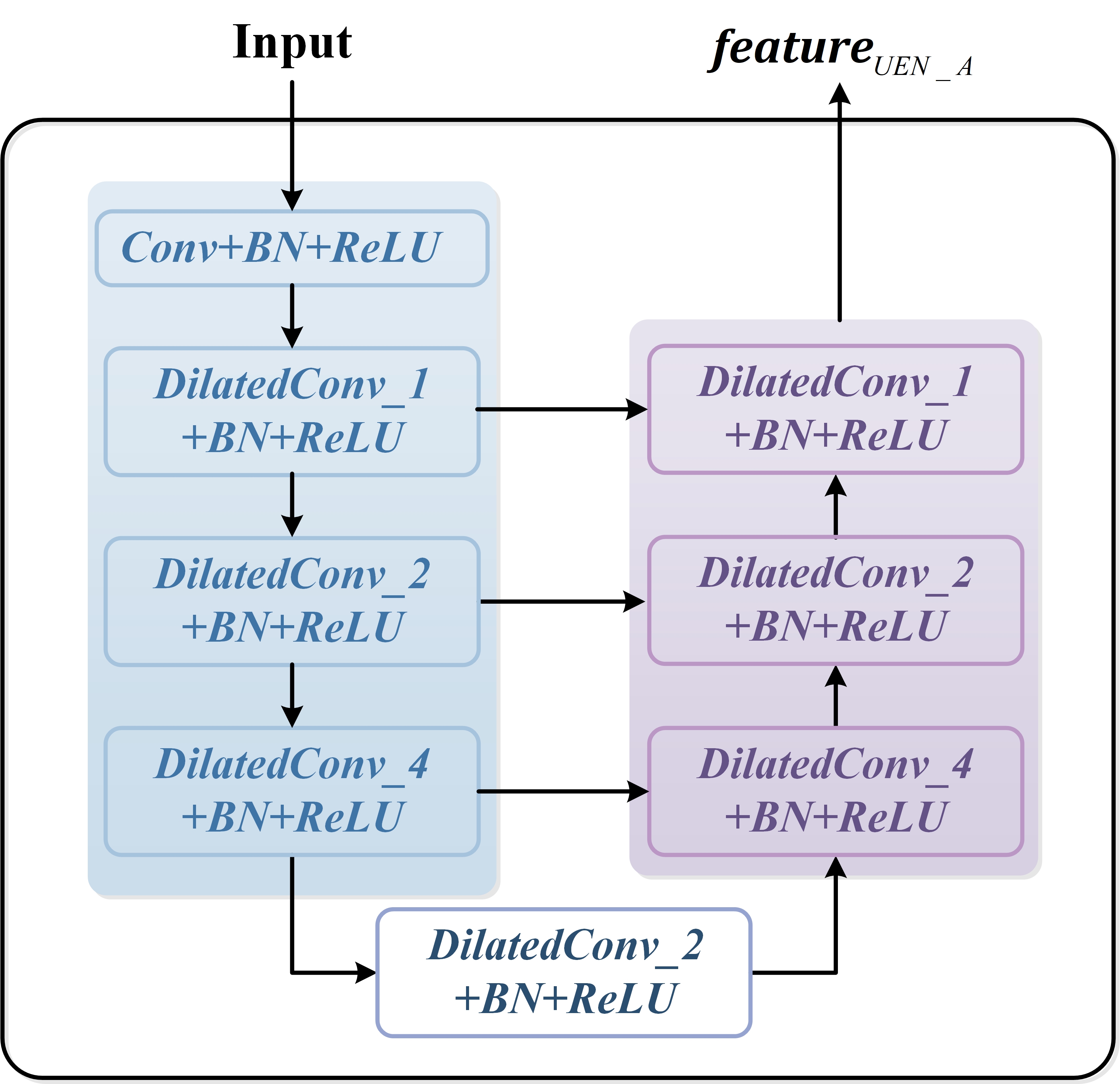}
	}
	\caption{The structures of UEN5 and UEN\_A.}
	\label{fig3}
\end{figure}
Fig. \ref{fig3} shows the overview of different UEN modules, including the $UEN\_5$ and the $UEN\_A$. $DilatedConv_i$ in Fig. \ref{fig3} denotes the atrous convolution layer with dilation rate $i$. The $3\times3$ convolutional layer is heavily used inside this module. Except for the first convolutional layer, each convolutional layer is followed by a down-sampling by a factor of two. In this way, richer information with a global view can be extracted at different scales. We determine the down-sampling times in an $UEN\_i$ $( i = 4,5)$ according to the scale of the input feature. If the input is large enough, such as $112\times112$, we could down-sample for three times. However, if the input itself is small, excessive down-sampling is redundant and unreasonable. In this situation, two down-sampling operations are enough. For smaller scales, such as $7\times7$, the module discards the down-sampling. More specifically, we remove the pooling layers and change the original convolutional layer into a dilated convolutional layer.
The structure shown in Fig. \ref{fig3} (a) is the specific settings of $UEN\_5$. Compared to $UEN\_5$, $UEN\_4$ reduces a down-sampling operation and an up-sampling operation followed by a convolutional layer, respectively. After the encoder, a dilated convolutional layer is followed by the batch normalization and ReLU activation function. Here, the input of each convolutional layer is an aggregation of two features. One is the output from the last convolution. The other one is the feature transmitted from the encoder. Here, the two features are aggregated by concatenating. 
Unlike $UEN\_i$, in Fig. \ref{fig3} (b), there are three dilated convolution branches in $UEN\_A$ with different dilation rates to extract semantic features with different receptive fields. Kernels of the three branches are all $3\times3$ and dilation rates are 1, 2, 4. The final convolution output $Feature_{UEN}$ is conducted as the output of this module.  The input size of each UEN block and the type used in each stage are present in Table. \ref{table1}. 
\begin{table*}[h]
	\centering
	\caption{Input size and the UEN version of each stage in backbone.}
	\setlength{\tabcolsep}{2.5mm}{
		\begin{tabular}{ccc}
			\hline  
			stage&Input size&The type of UEN\\
			\hline  
			Stage1&$112 \times 112$&$UEN\_5$\\
			Stage2&$56 \times 56$&$UEN\_4$\\
			Stage3&$28 \times 28$&$UEN\_4$\\
			Stage4&$14 \times 14$&$UEN\_A$\\
			Stage5&$7 \times 7$&$UEN\_A$\\
			\hline 
			\label{table1}
	\end{tabular}}
\end{table*}

\subsection{Additional Down-sampling Module}
\begin{figure*}[ht]
	\centering
	\includegraphics[width=12cm]{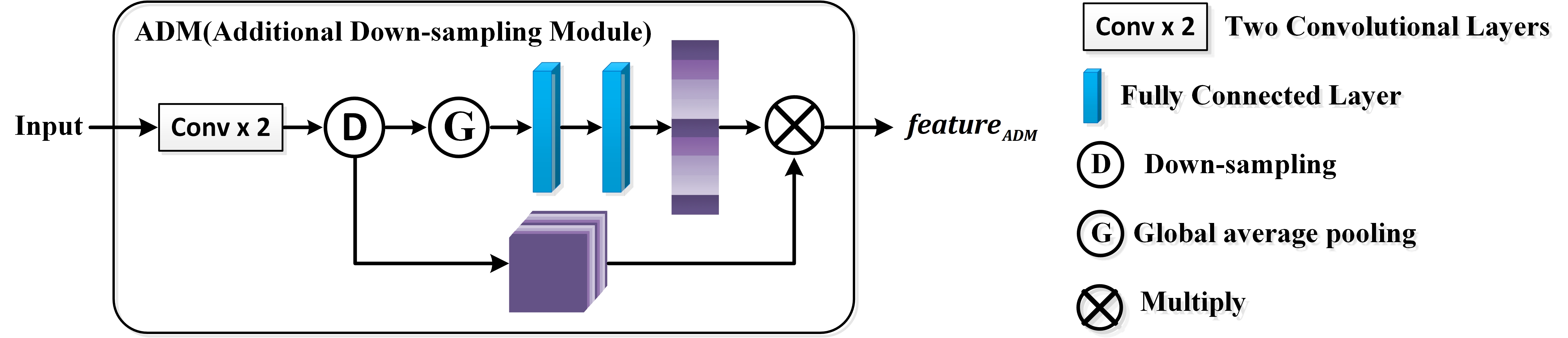}
	\caption{The structure of ADM.}
	\label{fig4}
\end{figure*}

In some scenarios of SOD, some salient areas are quite similar to the non-salient areas around them, for example, the salient object and the reflection of itself in the mirror. This problem leads to that sometimes models will treat these easily-confused un-salient regions as salient regions. The purpose of the Additional Down-sampling Module is to explore more global semantic clues as much as possible to assist the model position accurately and eliminate misleading regions. 
The structure of ADM is shown in Fig. \ref{fig4}. The input is down-sampled after two convolutional layers. It is reasonable to infer that the more non-zero pixels in a channel, the more information contained. On the contrary, if a channel has comparatively more zero pixels than other channels, we suppose the information content is inadequate. The global average pooling is followed by two fully connected layers to calculate the amount of information in each channel. Each element in the processed vector corresponds to a channel of the down-sampled feature. The down-sampled feature is multiplied by the vector in channel-wise to make a channel enhancement. As a result, the weights of these channels lacking information are weakened and the others are highlighted. The strengthened $Feature_{ADM}$ is used as the final output of ADM. 
\subsection{Loss Function}
Our algorithm supervises five outputs of the model, among which one output is the edge saliency map. Since there is no ground truth of the saliency object edge in the training set, it is necessary to generate the edge labels first. We calculate the gradient for each pixel of the original label. Then, multiply the gradient by itself to get the new value of each pixel:
\begin{equation}
	S_{edge}=d_{x}^{2}+d_{y}^{2}
\end{equation}
$S_{\text {edge }}$ is the edge ground truth wanted. $d_{x}$ and $d_{y}$ are the gradient values of the pixel in the x-direction and y-direction, respectively. Finally, the non-zero pixels are set as one and then multiply by 255. The edge saliency map is supervised by the edge ground truth generated in this way. The edge loss inspired by \cite{12WFNet2020} is written as:
\begin{equation}
	Edge\_Loss=-\sum_{e\in E^{+}} \log Pr\left(e_{n}=1\mid W\right)-\sum_{e\in E^{-}} \log Pr\left(e_{n}=0\mid W\right)
\end{equation}
$e_{n}$ represents the pixels in the edge prediction map, and $W$ represents the parameters in the model. $\operatorname{Pr}\left(e_{n}=1 \mid W\right)$ means the probability that the pixel $e_{n}$ is calculated as salient edge. $E^{+}$ indicates the salient edge pixel set, while $E^{-}$ is a set that contains all of the pixels except salient edge pixels.
Original labels supervise those intermediate saliency maps. For better performance, this method introduces two kinds of loss functions: $BCE\_Loss$ and $IoU\_Loss$. 
BCE Loss is widely used in the training process of binary classifiers. The function is written as:
\begin{equation}
	BCE\_{Loss}=-\sum_{(x,y)}{[g(x,y)\log p(x,y)+(1-g(x,y)\log (1-p(x,y)))]},
\end{equation}
where $g(x, y)\in[0, 1]$ is the ground truth label of the pixel
$(x,y)$ and $p(x,y)\in[0,1]$ is the predicted probability of
being salient object.
IoU Loss is joint to highlight the prediction error around the salient object boundary. As mentioned before, one of the difficulties in saliency object detection is that the prediction error frequently occurs in the object boundary area, which makes the prediction edge ambiguous and leads to error and missed detection. If the saliency map fails to align with the ground truth, the unaligned part will pay the penalty. The addition of IoU Loss can help the model to correct errors in this specified area. The function is denoted as
\begin{equation}
	sum_{inter}=\sum_{y_{i} \in I}y_{i}, \quad I=Sal\_map\odot Ground\_Truth,
\end{equation}
\begin{equation}
	 sum_{union}=\sum_{y_{j} \in J}y_{j}, \quad J=Sal\_map+Ground\_Truth,
\end{equation}
\begin{equation}
	IoU\_Loss={\left(1-\frac{sum_{inter}+1}{sum_{union}-sum_{inter}+1}\right)}\times \frac{1}{H \times W}
\end{equation}
in which $H$ is the height of $Sal\_map$ while $W$ is the width.
In the experiment, we found that the lower resolution of a saliency map, the more inaccurate the result is. If the losses of all saliency map are given the same weight, it will lead to instability and influence the model performance. Saliency maps generated from deep layers should be weakened with lower weight, so the coefficient is reduced by two. The whole loss function could be written as
\begin{equation}
	\begin{aligned}
		L_{a l l}= L_{edge}+B C E_{-} Loss_{\text {united }}+ IoU\_Loss_{united} \\
				  +\sum_{i=1}^{5} \frac{1}{2^{i-1}}({B C E\_Loss }_{i}+IoU\_Loss_{i})
	\end{aligned}
\end{equation}
 $BCE_{-} Loss_{united}+ IoU\_Loss_{united}$ is the loss of united saliency maps.  $BCE_{-} Loss_{i}+ IoU\_Loss_{i}$ is the loss of the $i$th intermediate saliency map. The lager $i$ is, the closer the saliency map is to the bottom.

\section{Experiment}
\subsection{Experimental Setting}
\textbf{Datasets and implementation details.} The model is trained on the DUTS-TR with 10553 images. In detail, we trained the model using the SGD optimizer with initial learning rate 3e-5, 0.9 momentum, 5e-4 weight decay, and batch size 16.
Because the ResNet-50 parameters are pre-trained on ImageNet, the learning rate of this part is a tenth of the randomly initialized parts which is set as 3e-5.  Then, the trained model is tested on five datasets, including DUTS-TE with 5019 images, DUT-OMROM with 5168 images, HKU-IS with 4447 images, ECSSD with 1000 images and PASCAL-S with 850 images. 

\textbf{Metrics.} We comprehensively evaluate the model with four metrics widely used in SOD, including mean F-measure ($mF$), mean absolute error ($MAE$), structure-measure ($Sm$) and enhanced-alignment measure ($Em$). Mean F-measure is the average of F-measures, which is calculated as:
\begin{equation}
	F_{\beta}=\frac{\left(1+\beta^{2}\right)\times{Precision} \times{Recall}}{\beta^{2} \times{Precision}+{Recall}}
\end{equation}
$\beta^{2}$ is usually set as 0.3 to put more emphasis on precision \cite{23Is2004}. Mean absolute error is the average of each pixel’s absolute error in predictions. Before calculation, each pixel is normalized to $[0,1]$, and $MAE$ is calculated as
\begin{equation}
	MAE=\frac{1}{H  \times W} \sum_{i=1}^{H} \sum_{j=1}^{W}|P(i, j)-G(i, j)|
\end{equation}
$H$ and $W$ is the height and width of the saliency map. $P(i, j)$ is the prediction at location $(i, j)$. $G(i, j)$ denotes the value at location $(i, j)$ in the ground truth. S-measure is a metric proposed to evaluate the structural similarity between the prediction and the corresponding ground truth. S-measure is formulated as
\begin{equation}
	S=\alpha S_{o}+(1-\alpha) S_{r},
\end{equation}
where $S_{o}$ represents the object-oriented structural similarity while $S_{r}$ represents the region-oriented structural similarity. $\alpha$ is set to 0.5, the same as in \cite{24Structure2017}. E-measure \cite{25Enhanced2018} is widely used in SOD as well. 
\begin{equation}
	E=\frac{1}{W\times H} \sum_{x=1}^{w} \sum_{y=1}^{h} \phi FM(i, j)
\end{equation}
where $\phi FM(i, j)$ is the enhanced alignment matrix. E-measure considers the performance of both the global average of image and local pixel when evaluating the model.

\subsection{Ablation Study}
\begin{figure*}[ht]
	\centering
	\includegraphics[width=12cm]{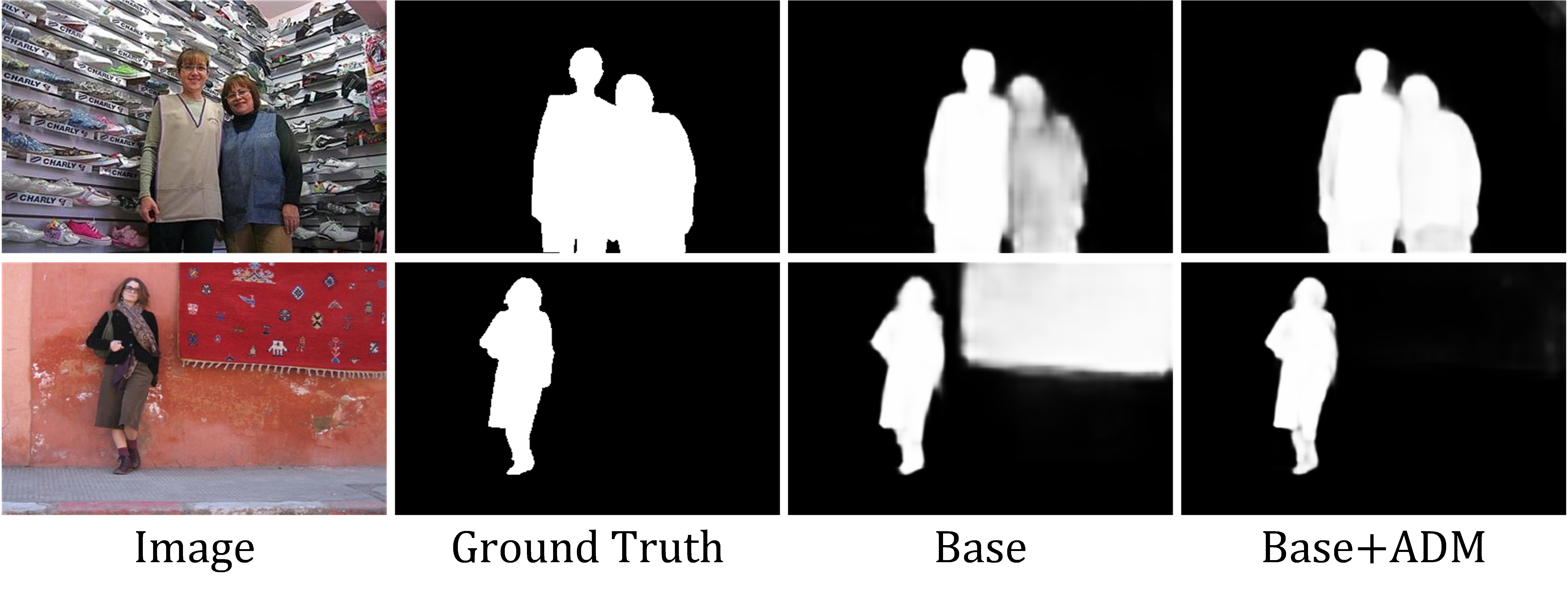}
	\caption{Comparison of predictions between the vanilla network with and without ADM.}
	\label{fig5}
\end{figure*}

In general, there is no need to down-sample the feature at $7 \times 7$ scale because it has a sufficiently large receptive field. However, the experiment results in Table. \ref{tab2} is somewhat counterintuitive. It shows that there is still some different global semantic information waiting to be discovered. The version for comparison is a U-shaped vanilla model, which is trained with the initial learning rate $10^{-4}$. Here we use the same learning rate to train the parameters in the backbone and modules designed in our method. The rise of $MAE$, $mF$, $Sm$ and $Em$ reflects ADM's reliability and validity. Fig. \ref{fig5} shows the comparison of visualized results between the vanilla network with and without ADM. UENs generate multi-scale features on account of the input level and integrate the features with edge information, which guides the model to pay attention to the boundary areas. The experimental data also reveals that the UENs make greater improvements in various aspects. 
\begin{table*}[h]
	\centering
	\caption{Comparison of networks with and without the proposed modules, ADM and UENs. Base: a U-structure network with usual U-shape modules. }
	\setlength{\tabcolsep}{1.1mm}{
		\begin{tabular}{ccc|cccc|cccc}
			\hline  
			\multicolumn{3}{c|}{Datasets}&\multicolumn{4}{c|}{DUTS-TE}&\multicolumn{4}{c}{DUT-OMROM}\\
			\hline  
			Base&+ADM&+UENs&$mF\uparrow$&$MAE\downarrow$&$Sm\uparrow$&$Em\uparrow$&$mF\uparrow$&$MAE\downarrow$&$Sm\uparrow$&$Em\uparrow$\\
			\hline
		    \checkmark&&&0.851&0.038&0.894&0.901&0.772&0.065&0.839&0.859\\
			\checkmark&\checkmark&&0.860&0.035&0.898&0.907&0.778&0.059&0.843&0.870\\
			\checkmark&\checkmark&\checkmark&0.870&0.031&0.904&0.917&0.790&0.052&0.851&0.881\\
			\hline 
	\end{tabular}}
	\label{tab2}
\end{table*}

\subsection{Comparison with Other State-of-the-art Methods}
\begin{figure*}[ht]
	\centering
	\includegraphics[width=12.5cm]{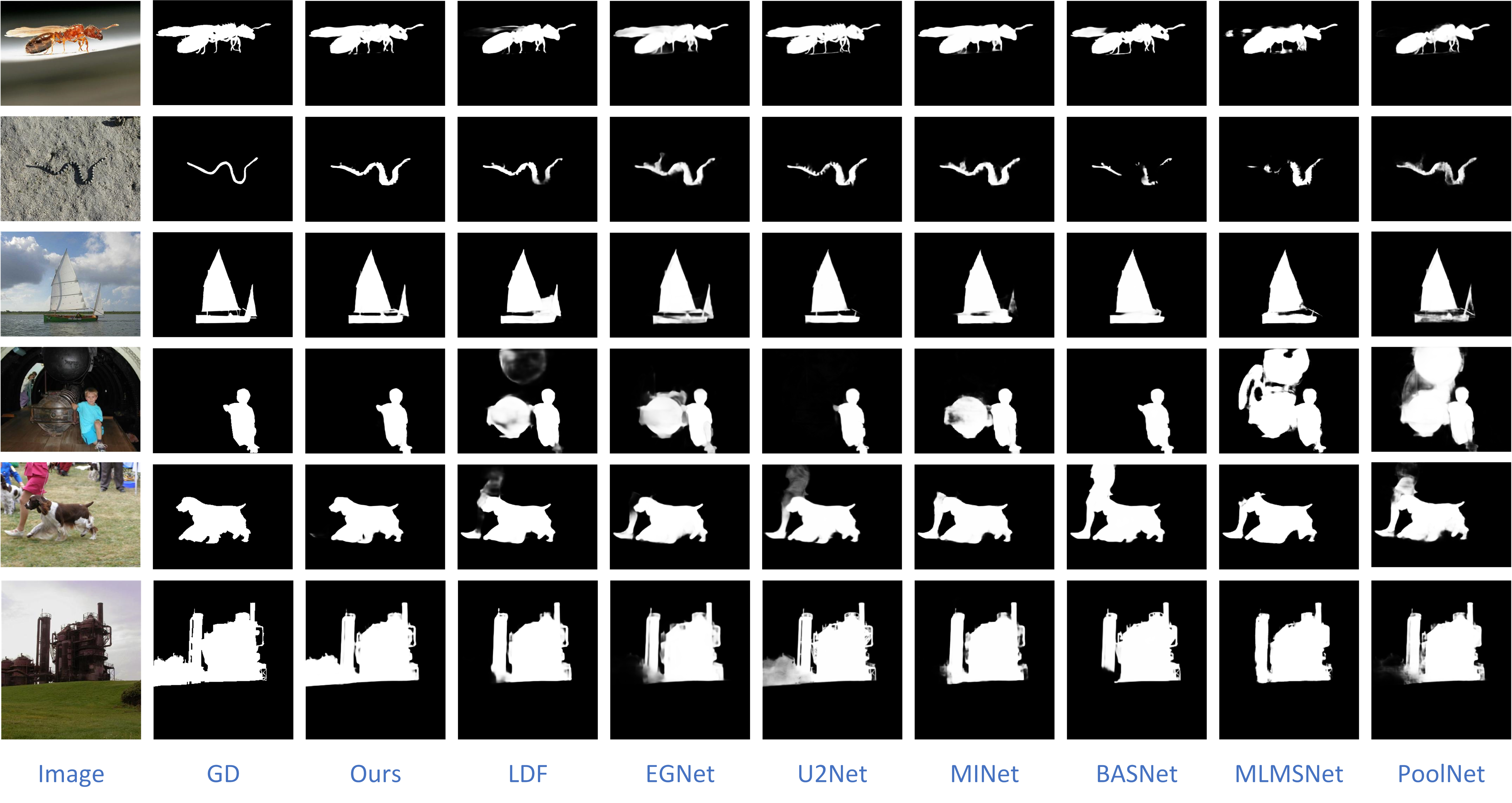}
	\caption{Visual comparison of the previous SOTA methods and our method denoted as 'Ours'.}
	\label{fig6}
\end{figure*}

A large number of experiments are conducted to convince the validity of our approach. To make a clear comparison with other state-of-the-art (SOTA) methods, we list all the results in Table. \ref{table4}. There are 15 SOTA methods proposed in the past three years, namely, BMPM \cite{35BMPM2018}, RAS \cite{36RAS2018}, PiCANet \cite{38PiCANet2018}, R3Net \cite{13R3Net2018}, BASNet \cite{39BASNet2019}, PoolNet \cite{40PoolNet2019}, TDBU \cite{41TDBU2019}, PAGE \cite{42PAGE2019},F3Net \cite{43F3Net2019}, CPD \cite{44CPD2019}, EGNet \cite{11EGNet2019}, MLMSNet \cite{45MLMSNet2020}, U2Net \cite{10U2Net2020}, MINet \cite{32MINet2020}, LDF \cite{46LDF2020}. For a fair comparison, all of the compared saliency maps are provided or generated from released models by the authors. The evaluation codes are all the same. As shown in Table. \ref{table4} and Fig. \ref{fig6}, our algorithm surpasses most of the methods.

\begin{table*}[htp!]
	\centering
	\caption{Quantitative comparison with state-of-the-art methods on five datasets. The best results are highlighted in bold. The best and the second best results are highlighted in red and green respectively.}
	\setlength{\tabcolsep}{0.2mm}{
		\begin{tabular}{p{20mm}|cccc|cccc|cccc|cccc}
			\hline  
			Datasets&\multicolumn{4}{c|}{DUTS-TE}&\multicolumn{4}{c|}{ECSSD}&\multicolumn{4}{c|}{DUT-OMROM}&\multicolumn{4}{c}{HKU-IS}\\
			\hline
			Metrics&$mF$&MAE&$Sm$&$Em$&$mF$&MAE&$Sm$&$Em$&$mF$&MAE&$Sm$&$Em$&$mF$&MAE&$Sm$&$Em$\\
			\hline  
			BMPM\cite{35BMPM2018}&.745&.049&.862&.860&.868&.044&.911&.914&.692&.064&.809&.837&.871&.038&.907&.937\\
		    RAS\cite{36RAS2018}&.751&.059&.839&.861&.889&.059&.893&.914&.713&.062&.814&.846&.874&.045&.888&.931\\
			R3Net\cite{13R3Net2018}&.785&.057&.834&.867&.914&.040&.910&{\color{green}{.929}}&.747&.063&.815&.850&.893&.036&.895&.939\\
			PiCANet\cite{38PiCANet2018}&.749&.051&.867&.852&.885&.044&.917&.910&.710&.065&.835&.834&.870&.039&.908&.934\\
			MLMSNet\cite{45MLMSNet2020}&.799&.045&.856&.882&.914&.038&.911&.925&.735&.056&.817&.846&.892&.034&.901&.945\\
			PAGE\cite{42PAGE2019}&.777&.051&.854&.869&.906&.042&.912&.920&.736&.066&.824&.853&.882&.037&.903&.940\\	
			CPD\cite{44CPD2019}&.805&.043&.869&.886&.917&.037&.918&.925&.747&.056&.825&.866&.891&.034&.905&.944\\
			BASNet\cite{39BASNet2019}&.756&.048&.866&.884&.880&.037&.916&.921&.756&.056&.836&.869&.895&.032&.909&.946\\
			F3Net\cite{43F3Net2019}&.840&.035&.888&.902&.925&.033&.924&.927&.766&.053&.838&.870&.840&.062&.855&.859\\
			PoolNet\cite{40PoolNet2019}&.799&.040&.879&.881&.910&.042&.917&.921&.739&.055&.832&.858&.885&.032&-&.941\\
			TDBU\cite{41TDBU2019}&.767&.048&.865&.879&.880&.040&.918&.922&.739&.059&.837&.854&.878&.038&.907&.942\\
			EGNet\cite{11EGNet2019}&.815&.039&.875&.891&.920&.041&.918&.927&.755&{\color{green}{.052}}&.818&.867&.898&.031&.918&.948\\
			U2Net\cite{10U2Net2020}&.792&.044&.861&.886&.892&{\color{green}{.033}}&{\color{green}{.928}}&.924&.761&.054&{\color{green}{.847}}&.871&.896&.031&.916&.948\\
			MINet\cite{32MINet2020}&.828&.037&.884&{\color{red}{.917}}&.924&{\color{green}{.033}}&.925&{\color{red}{.953}}&.756&.055&.833&{\color{green}{.873}}&.908&.028&{\color{green}{.920}}&{\color{red}{.961}}\\
			LDF\cite{46LDF2020}&{\color{green}{.855}}&{\color{green}{.034}}&{\color{green}{.892}}&{\color{green}{.910}}&{\color{green}{.930}}&.034&.924&.925&{\color{green}{.773}}&{\color{red}{.051}}&.838&{\color{green}{.873}}&{\color{green}{.914}}&{\color{green}{.027}}&.919&.954\\
			\hline
			Ours&{\color{red}{.870}}&{\color{red}{.031}}&{\color{red}{.904}}&{\color{red}{.917}}&{\color{red}{.936}}&{\color{red}{.028}}&{\color{red}{.934}}&{\color{green}{.929}}&{\color{red}{.790}}&{\color{green}{.052}}&{\color{red}{.851}}&{\color{red}{.881}}&{\color{red}{.917}}&{\color{red}{.026}}&{\color{red}{.925}}&{\color{green}{.956}}\\
			\hline 
	\end{tabular}}
	\label{table4}
\end{table*}

\textbf{Quantitative comparison.} Table. \ref{table4} shows the quantitative evaluation results of the SOTA methods mentioned above and our model in terms of $mF$, $MAE$, $Sm$, and $Em$. The proposed method consistently performs better than all the competitors across four metrics on four datasets. In terms of $Em$, our method achieves the second best overall performance, which is slightly inferior to MINet. It is worth noting that MEUNet achieves the best performance in terms of the mean F-measure and structure quality evaluation $Sm$.

\textbf{Qualitative comparison.} Fig. \ref{fig6} shows the visual comparison between our model and other SOTA methods. The first column is the images and the second column is the corresponding ground truths. Our result is in the third column. It can be observed that our method could locate the salient object accurately and segment the foreground and background around the boundary areas precisely. 
Predictions in rows 1, 2 and 3 reveal the abilities of the models to deal with detailed extraction, and other lines indicate the location ability. The method proposed in this article is good at keeping as much detail as possible with sharp edges.
For example, the legs of the bee in the first picture are clearly present in its prediction, and the snake is continuous and complete only on our map. We suppose that the detail retention ability ascribes to the UENs and the location ability owes to the ADM. In addition to this, the artful structural design allows the features from different layers to complement each other and fuse properly. Our model achieves the best in terms of the overall effect.

\section{Conclusion}
In this paper, we present a Multi-scale Edge-based U-shape Network with both enhanced high-level semantic features and low-level detail information for SOD. The central architecture of our network is a U-shape encoder-decoder structure which mainly consists of UENs and ADM. The UEN is an outstanding evolution of U-Net, and it contains multi-scale features with rich information and edge information that could assist boundary prediction. ADM can help the network to discover semantic clues from a more global view. Besides, we employ some tricks to improve the performance of the model. For example, the edge saliency map is supervised by salient edges to suppress the edges from the background. The fusion of the parallel four saliency maps is outputed as the final result.

\bibliographystyle{splncs04}
\bibliography{refers}

\begin{thebibliography}{10}
\providecommand{\url}[1]{\texttt{#1}}
\providecommand{\urlprefix}{URL }
\providecommand{\doi}[1]{https://doi.org/#1}

\bibitem{3Adaptive2012}
Borji, A., Frintrop, S., Sihite, D.N., Itti, L.: Adaptive object tracking by
  learning background context. In: 2012 IEEE Computer Society Conference on
  Computer Vision and Pattern Recognition Workshops. pp. 23--30. IEEE (2012)

\bibitem{12WFNet2020}
Cen, J., Sun, H., Chen, X., Liu, N., Liang, D., Zhou, H.: Wfnet: A wider and
  finer network for salient object detection. IEEE Access  \textbf{8},
  210418--210428 (2020)

\bibitem{36RAS2018}
Chen, S., Tan, X., Wang, B., Hu, X.: Reverse attention for salient object
  detection. In: Proceedings of the European Conference on Computer Vision
  (ECCV). pp. 234--250 (2018)

\bibitem{14Contour2021}
Chen, Z., Zhou, H., Lai, J., Yang, L., Xie, X.: Contour-aware loss:
  Boundary-aware learning for salient object segmentation. IEEE Transactions on
  Image Processing  \textbf{30},  431--443 (2020)

\bibitem{13R3Net2018}
Deng, Z., Hu, X., Zhu, L., Xu, X., Qin, J., Han, G., Heng, P.A.: R³net:
  Recurrent residual refinement network for saliency detection. In: Proceedings
  of the 27th International Joint Conference on Artificial Intelligence. pp.
  684--690 (7 2018)

\bibitem{24Structure2017}
Fan, D.P., Cheng, M.M., Liu, Y., Li, T., Borji, A.: Structure-measure: A new
  way to evaluate foreground maps. In: Proceedings of the IEEE international
  conference on computer vision. pp. 4548--4557 (2017)

\bibitem{25Enhanced2018}
Fan, D.P., Gong, C., Cao, Y., Ren, B., Cheng, M.M., Borji, A.:
  Enhanced-alignment measure for binary foreground map evaluation. arXiv
  preprint arXiv:1805.10421  (2018)

\bibitem{31Residual2020}
Feng, M., Lu, H., Yu, Y.: Residual learning for salient object detection. IEEE
  Transactions on Image Processing  \textbf{29},  4696--4708 (2020)

\bibitem{22Deep2016}
He, K., Zhang, X., Ren, S., Sun, J.: Deep residual learning for image
  recognition. In: Proceedings of the IEEE conference on computer vision and
  pattern recognition. pp. 770--778 (2016)

\bibitem{33Learning2020}
Li, T., Song, H., Zhang, K., Liu, Q.: Learning residual refinement network with
  semantic context representation for real-time saliency object detection.
  Pattern Recognition  \textbf{105},  107372 (2020)

\bibitem{30HFFNet2020}
Li, X., Song, D., Dong, Y.: Hierarchical feature fusion network for salient
  object detection. IEEE Transactions on Image Processing  \textbf{29},
  9165--9175 (2020)

\bibitem{40PoolNet2019}
Liu, J., Hou, Q., Cheng, M.M., Feng, J., Jiang, J.: A simple pooling-based
  design for real-time salient object detection. 2019 IEEE/CVF Conference on
  Computer Vision and Pattern Recognition (CVPR) pp. 3912--3921 (2019)

\bibitem{38PiCANet2018}
Liu, N., Han, J., Yang, M.H.: Picanet: Learning pixel-wise contextual attention
  for saliency detection. In: Proceedings of the IEEE Conference on Computer
  Vision and Pattern Recognition. pp. 3089--3098 (2018)

\bibitem{9Fully2015}
Long, J., Shelhamer, E., Darrell, T.: Fully convolutional networks for semantic
  segmentation. In: Proceedings of the IEEE conference on computer vision and
  pattern recognition. pp. 3431--3440 (2015)

\bibitem{1An2002}
Ma, Y.F., Lu, L., Zhang, H.J., Li, M.: A user attention model for video
  summarization. In: Proceedings of the tenth ACM international conference on
  Multimedia. pp. 533--542 (2002)

\bibitem{32MINet2020}
Pang, Y., Zhao, X., Zhang, L., Lu, H.: Multi-scale interactive network for
  salient object detection. In: Proceedings of the IEEE/CVF Conference on
  Computer Vision and Pattern Recognition. pp. 9413--9422 (2020)

\bibitem{7Saliency2012}
Perazzi, F., Kr{\"a}henb{\"u}hl, P., Pritch, Y., Hornung, A.: Saliency filters:
  Contrast based filtering for salient region detection. In: 2012 IEEE
  conference on computer vision and pattern recognition. pp. 733--740. IEEE
  (2012)

\bibitem{10U2Net2020}
Qin, X., Zhang, Z., Huang, C., Dehghan, M., Zaiane, O.R., Jagersand, M.:
  U2-net: Going deeper with nested u-structure for salient object detection.
  Pattern Recognition  \textbf{106},  107404 (2020)

\bibitem{39BASNet2019}
Qin, X., Zhang, Z., Huang, C., Gao, C., Dehghan, M., Jagersand, M.: Basnet:
  Boundary-aware salient object detection. In: Proceedings of the IEEE/CVF
  Conference on Computer Vision and Pattern Recognition. pp. 7479--7489 (2019)

\bibitem{26UNet2015}
Ronneberger, O., Fischer, P., Brox, T.: U-net: Convolutional networks for
  biomedical image segmentation. In: International Conference on Medical image
  computing and computer-assisted intervention. pp. 234--241. Springer (2015)

\bibitem{23Is2004}
Rutishauser, U., Walther, D., Koch, C., Perona, P.: Is bottom-up attention
  useful for object recognition? In: Proceedings of the 2004 IEEE Computer
  Society Conference on Computer Vision and Pattern Recognition, 2004. CVPR
  2004. vol.~2, pp. II--II. IEEE (2004)

\bibitem{41TDBU2019}
Wang, W., Shen, J., Cheng, M.M., Shao, L.: An iterative and cooperative
  top-down and bottom-up inference network for salient object detection. In:
  Proceedings of the IEEE/CVF Conference on Computer Vision and Pattern
  Recognition. pp. 5968--5977 (2019)

\bibitem{42PAGE2019}
Wang, W., Zhao, S., Shen, J., Hoi, S., Borji, A.: Salient object detection with
  pyramid attention and salient edges. 2019 IEEE/CVF Conference on Computer
  Vision and Pattern Recognition (CVPR) pp. 1448--1457 (2019)

\bibitem{43F3Net2019}
Wei, J., Wang, S., Huang, Q.: F³net: Fusion, feedback and focus for salient
  object detection. In: AAAI (2020)

\bibitem{46LDF2020}
Wei, J., Wang, S., Wu, Z., Su, C., Huang, Q., Tian, Q.: Label decoupling
  framework for salient object detection. 2020 IEEE/CVF Conference on Computer
  Vision and Pattern Recognition (CVPR) pp. 13022--13031 (2020)

\bibitem{45MLMSNet2020}
Wu, R., Feng, M., Guan, W., Wang, D., Lu, H., Ding, E.: A mutual learning
  method for salient object detection with intertwined multi-supervision. 2019
  IEEE/CVF Conference on Computer Vision and Pattern Recognition (CVPR) pp.
  8142--8151 (2019)

\bibitem{15EDN2020}
Wu, Y., Liu, Y., Zhang, L., Cheng, M.M.: Edn: Salient object detection via
  extremely-downsampled network. ArXiv  \textbf{abs/2012.13093} (2020)

\bibitem{44CPD2019}
Wu, Z., Su, L., Huang, Q.: Cascaded partial decoder for fast and accurate
  salient object detection. 2019 IEEE/CVF Conference on Computer Vision and
  Pattern Recognition (CVPR) pp. 3902--3911 (2019)

\bibitem{5Joint2019}
Zeng, Y., Zhuge, Y., Lu, H., Zhang, L.: Joint learning of saliency detection
  and weakly supervised semantic segmentation. In: Proceedings of the IEEE/CVF
  International Conference on Computer Vision. pp. 7223--7233 (2019)

\bibitem{35BMPM2018}
Zhang, L., Dai, J., Lu, H., He, Y., Wang, G.: A bi-directional message passing
  model for salient object detection. In: Proceedings of the IEEE Conference on
  Computer Vision and Pattern Recognition. pp. 1741--1750 (2018)

\bibitem{11EGNet2019}
Zhao, J.X., Liu, J.J., Fan, D.P., Cao, Y., Yang, J., Cheng, M.M.: Egnet: Edge
  guidance network for salient object detection. In: Proceedings of the
  IEEE/CVF International Conference on Computer Vision. pp. 8779--8788 (2019)

\bibitem{29Hierarchical2021}
Zhou, S., Wang, J., Wang, L., Zhang, J., Wang, F., Huang, D., Zheng, N.:
  Hierarchical and interactive refinement network for edge-preserving salient
  object detection. IEEE Transactions on Image Processing  \textbf{30},  1--14
  (2020)

\bibitem{27HUANet2020}
Zhou, S., Wang, J., Zhang, J., Wang, L., Huang, D., Du, S., Zheng, N.:
  Hierarchical u-shape attention network for salient object detection. IEEE
  Transactions on Image Processing  \textbf{29},  8417--8428 (2020)

\end{thebibliography}
%




\end{document}